\documentclass[conference]{IEEEtran}
\IEEEoverridecommandlockouts
\usepackage{cite}
\usepackage{amsmath,amssymb,amsfonts}
\usepackage{algorithmic}
\usepackage{graphicx}
\usepackage{textcomp}
\usepackage{xcolor}
\usepackage{hyperref}
\usepackage{multirow}
\def\BibTeX{{\rm B\kern-.05em{\sc i\kern-.025em b}\kern-.08em
    T\kern-.1667em\lower.7ex\hbox{E}\kern-.125emX}}
\begin{document}

\title{Stress-Query and Temporal Reliability of Vision-Language Models as copilots in Autonomous UAV Navigation
}

\author{\IEEEauthorblockN{1\textsuperscript{st} Hiago Sodre}
\IEEEauthorblockA{\textit{Robotics and AI Laboratory} \\
\textit{Technological University of Uruguay}\\
Rivera, Uruguay\\
hiago.sodre@utec.edu.uy}
\and
\IEEEauthorblockN{2\textsuperscript{rd} Sebastian Barcelona}
\IEEEauthorblockA{\textit{Robotics and AI Laboratory} \\
\textit{Technological University of Uruguay}\\
Rivera, Uruguay\\
sebastian.barcelona@utec.edu.uy}
\and
\IEEEauthorblockN{3\textsuperscript{rd} Vincent Sandin}
\IEEEauthorblockA{\textit{Robotics and AI Laboratory} \\
\textit{Technological University of Uruguay}\\
Rivera, Uruguay \\
vincent.sandin@estudiantes.utec.edu.uy}
\and
\IEEEauthorblockN{4\textsuperscript{th} Pablo Moraes}
\IEEEauthorblockA{\textit{Robotics and AI Laboratory} \\
\textit{Technological University of Uruguay}\\
Rivera, Uruguay \\
pablo.moraes@utec.edu.uy}
\and
\IEEEauthorblockN{5\textsuperscript{th} André S. Kelbouscas}
\IEEEauthorblockA{\textit{Robotics and AI Laboratory} \\
\textit{Technological University of Uruguay}\\
Rivera, Uruguay \\
andre.dasilva@utec.edu.uy}
\and
\IEEEauthorblockN{6\textsuperscript{th} Ricardo B. Grando}
\IEEEauthorblockA{\textit{Robotics and AI Laboratory} \\
\textit{Technological University of Uruguay}\\
Rivera, Uruguay \\
ricardo.bedin@utec.edu.uy}
}

\maketitle

\begin{abstract}
Integrating Vision-Language Models (VLMs) into autonomous UAV navigation enables high-level semantic reasoning but introduces critical temporal bottlenecks that can destabilize flight control. This paper presents a systematic benchmark that evaluates the operational trade-offs among model scale, hardware constraints, and flight reliability. By subjecting the system to extensive stress tests on both a high-performance workstation and a constrained integrated-graphics laptop, we isolate the kinetic impact of inference latency against reasoning accuracy. Our findings reveal a performance dichotomy: lightweight models struggle with structural hallucinations that break deterministic state machines, while heavyweight models induce trajectory overcorrections due to inference lag. We demonstrate that a mid-weight architecture strikes the optimal balance, successfully rehabilitating navigation in degraded environments without compromising real-time stability across different hardware profiles. A video of the tests is available at: \url{https://youtu.be/vHtjc7HarJ0}
\end{abstract}

\begin{IEEEkeywords}
Autonomous Navigation, Unmanned Aerial Vehicles, Vision-Language Models,  Temporal Reliability.
\end{IEEEkeywords}

\section{Introduction}

Traditional autonomous navigation tasks in Unmanned Aerial Vehicles (UAVs) have relied on conventional perception pipelines that utilize rigid computer vision architectures and sensor fusion. While these geometric systems are highly reactive, their performance degrades significantly in uncertain environments, such as when navigation markers are severely damaged or visual cues are ambiguous. In contrast, the rapid development of Vision-Language Models (VLMs) represents a paradigm shift \cite{levine2016end}. By incorporating multimodal semantic reasoning into the perception process, UAVs can interpret high-level contextual information in complex scenarios. This approach bridges the gap between rigid pixel-level processing and contextual scene understanding, enabling UAVs to conduct a more intelligent secondary assessment of their surroundings.

However, the transition of VLMs from cloud-based APIs to real-time embedded robotic control presents profound operational challenges. The main obstacle is the bottleneck between model scale and inference time limits. Models with many parameters exhibit robust reasoning but impose inference latencies that often exceed several seconds on edge hardware, decoupling the control loop and causing severe deviations in the physical trajectory. On the other hand, lightweight edge-optimized models minimize latency but often suffer from structural distortions, failing to maintain the rigorous formatting schemes, such as JSON outputs, required by classical deterministic controllers. While previous research has explored end-to-end visuomotor policies \cite{levine2016end} or reinforcement learning \cite{kober2013reinforcement}, there is a critical gap in the empirical understanding of the exact time limits and trade-offs that dictate when a VLM mathematically breaks an autonomous control loop.

\begin{figure*}[h]
    \centering
    \includegraphics[width=1.70\columnwidth]{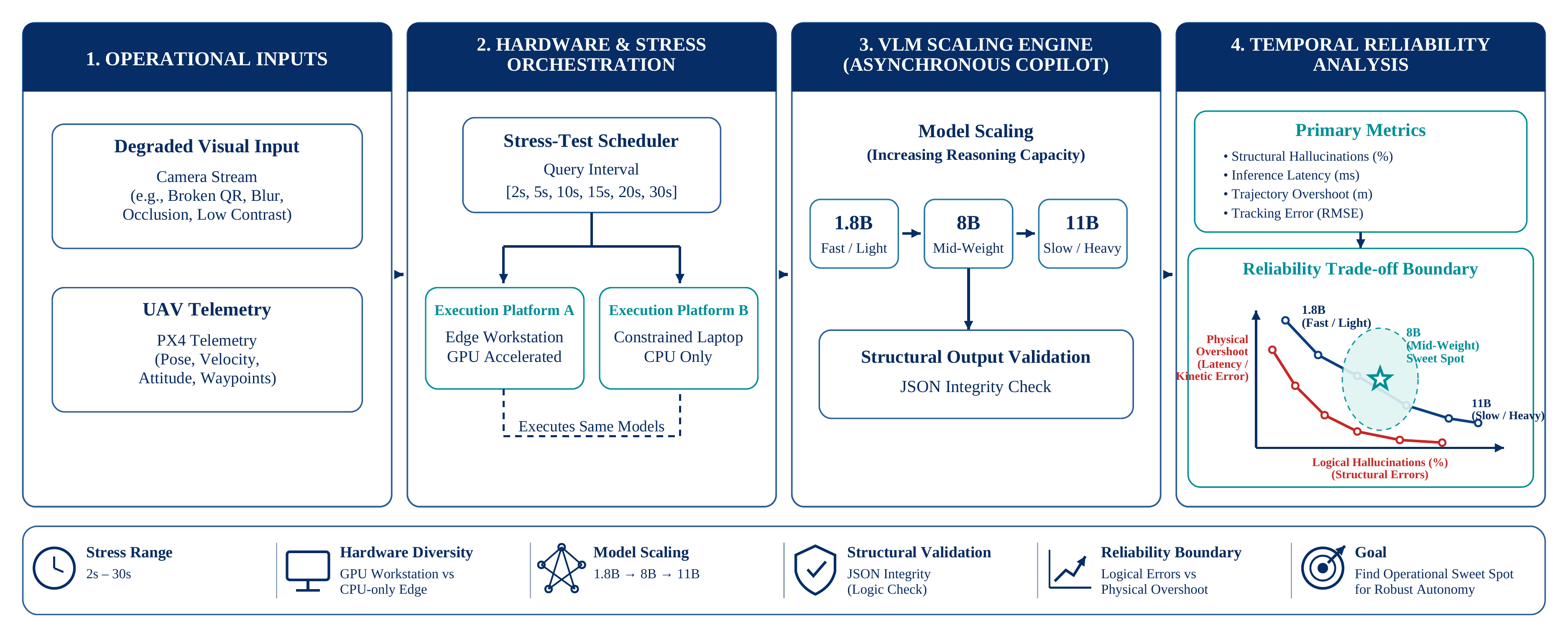}
    \caption{Proposed Hybrid FSM-VLM evaluation architecture. The deterministic FSM ensures strict real-time control, whilst the asynchronous VLM handles complex contextual anomalies.}
    \label{fig:architecture}
\end{figure*}

The evaluations are executed within a high-fidelity Software-In-The-Loop (SITL) Gazebo simulation, subjecting multiple visual architectures to incremental stress tests under both structured and severely degraded environments, as shown in the proposed diagram in Figure \ref{fig:architecture}.

The main contributions of this paper are summarized as follows:
\begin{itemize}
    \item A comprehensive quantitative benchmark that evaluates the real-time operational limits of varying VLM parameter scales (from 1.8B to 11B) in closed-loop UAV navigation systems natively available at \url{https://github.com/hiagojacobs/UAV_Sim_VLM/tree/stress}.
    \item The conceptualization and empirical measurement of the \textit{Temporal Reliability Boundary}, demonstrating how excessive hardware-induced latency directly correlates with physical trajectory overcorrections in real-time edge robotics.
    \item A systematic stress-testing protocol that varies asynchronous query intervals to isolate the critical trade-off between semantic reasoning capabilities and structural output hallucination rates in degraded environments.
    \item The identification of an operational "Sweet Spot" in mid-weight VLM architectures, establishing a blueprint for maximizing visual navigational recovery without compromising the strict timing requirements of deterministic FSM controllers.
\end{itemize}

\section{Related Works}

Reliable navigation in complex aerial missions requires the capacity to perceive, interpret, and respond to environmental information in real time. Traditional unmanned aerial vehicle (UAV) systems address this challenge using computer vision, sensor fusion, and deterministic control strategies, which provide robust performance in structured environments. However, their effectiveness declines significantly when visual cues are ambiguous, partially occluded, or degraded. These conditions reveal the inherent limitations of purely geometric perception methods, which restrict their ability to support advanced decision-making or navigational recovery \cite{chang2023gpsdenied}.

Recent advances in multimodal learning have introduced Vision-Language Models (VLMs) as a paradigm for incorporating abstract semantic reasoning into robotic perception systems. Models such as CLIP \cite{radford2021clip}, BLIP-2 \cite{li2023blip2}, and LLaVA demonstrate strong capabilities in visual understanding and zero-shot reasoning by integrating advanced visual encoders with large language models \cite{touvron2023llama}. Although these architectures have shown promising results on static, open-ended tasks, their application to dynamic, real-time autonomous systems is limited by significant inference latency and the lack of mathematical guarantees. 

To achieve both flexibility and deterministic safety in robotics, hybrid architectures that combine learning-based components with classical control systems have been extensively investigated. Previous research on visuomotor policies \cite{levine2016end} and reinforcement learning \cite{kober2013reinforcement} has demonstrated the effectiveness of data-driven approaches for sequential decision-making. Nevertheless, these methods generally function within narrowly defined action spaces and do not address the challenge of integrating large, free-form generative models into safety-critical, asynchronous control loops. Asynchronous control loops.

\section{System Overview and Benchmarking Framework}

\begin{figure}[h]
    \centering
    \includegraphics[width=0.83\columnwidth]{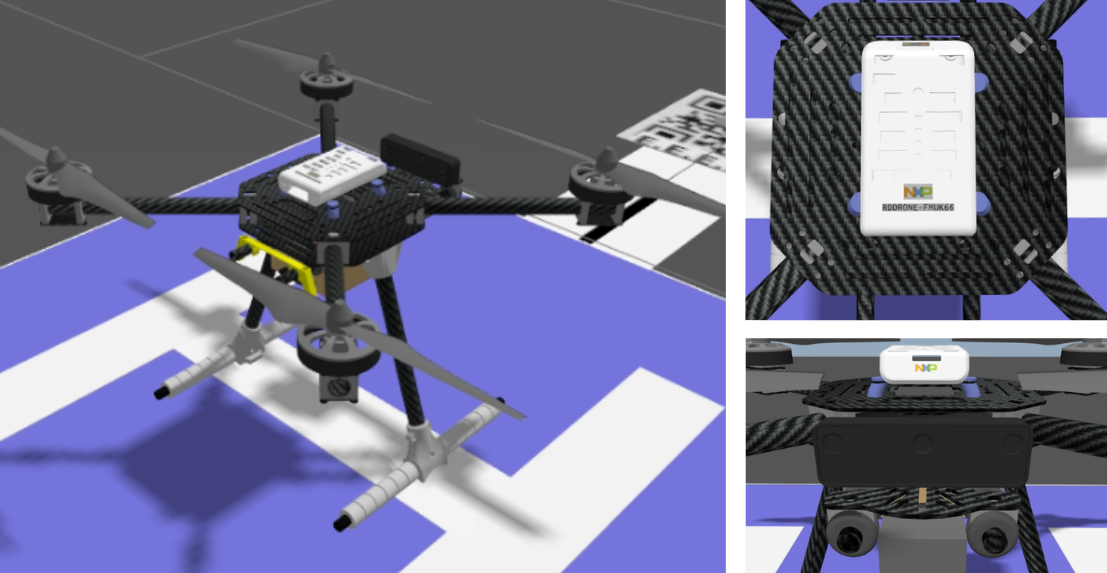}
    \caption{Autonomous Vehicle sensor payload and visual streams mapped within the Gazebo simulation environment.}
    \label{fig:frames}
\end{figure}

To systematically evaluate the impact of Vision-Language Model scaling and cognitive latency on real-time autonomous flight, an exhaustive benchmarking framework was developed. The evaluations are executed within a fully physics-accurate Software-In-The-Loop (SITL) stack running PX4 Autopilot, ROS~2 Humble middleware, and Gazebo Harmonic (Figure \ref{fig:frames}). 

\subsection{Hybrid Advisory Architecture}

\begin{figure}[h]
    \centering
    \includegraphics[width=0.99\columnwidth]{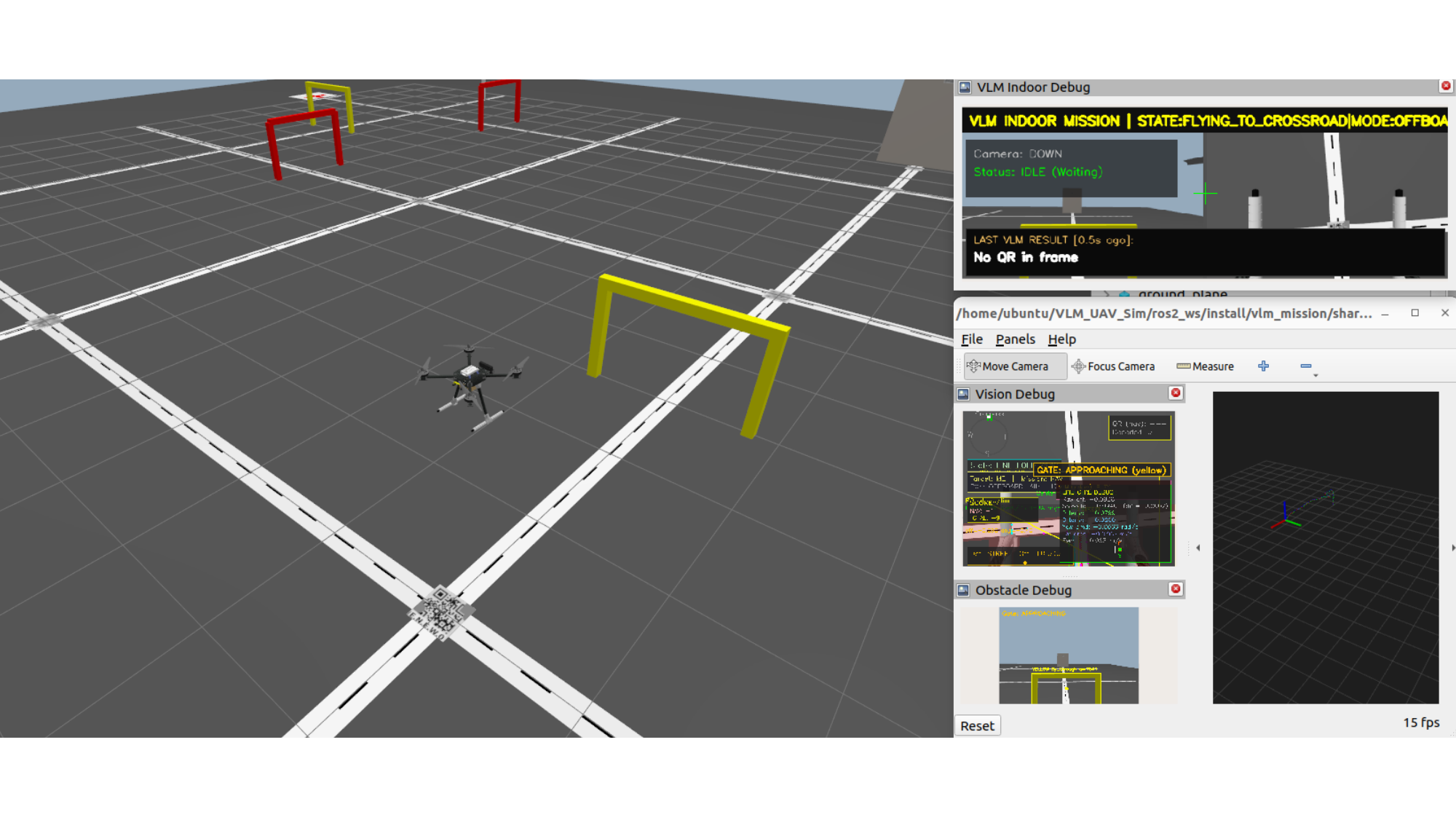}
    \caption{Simulation environment for the autonomous vehicle in an indoor area of the Gazebo, with the panels displaying RViz debugging overlays.}
    \label{fig:simulation_environment}
\end{figure}

The testbed operates under a decoupled hybrid control paradigm, deliberately separating low-level kinetic stability from high-level generative reasoning. 
The fast control layer consists of a deterministic Finite State Machine (FSM) operating at 30--50~Hz. This layer utilizes classical OpenCV-based pipelines to extract fast visual primitives (lines, simple geometric gates) and pushes immediate velocity setpoints to the flight controller via MAVROS, ensuring constant aerial stabilization how its represented by the Figure \ref{fig:simulation_environment}).

The slow generative layer functions as an asynchronous advisory copilot. Camera frames are periodically offloaded to a local VLM deployment running varying architectures via Ollama. By decoupling these layers, the FSM can maintain flight stability even while the VLM resolves queries that take several seconds. To combat desynchronization during these inference windows, the framework automatically triggers adaptive speed scaling (slowing the UAV geometry while awaiting insights) and records positional snapshots to enable Latency-Aware Backtracking if semantic target areas are breached during computation.

\subsection{Structured Reasoning Evaluation}

To quantify the reliability of generative models functioning as robotic operators, a strict structural contract was enforced. VLMs are queried through predefined spatial prompts and are required to output exclusively in a deterministic JSON schema (e.g., boolean flags for primitive detection, explicit strings for \textit{turn\_left}, and float values for \textit{confidence}). 

This interface allows the benchmark to mathematically isolate and track silent autonomous failures:
\begin{itemize}
    \item \textbf{Format Error}: The model returns unparsable syntax or conversational structures instead of the requested JSON.
    \item \textbf{Logic Error}: The response maintains valid syntax but hallucinates parameters or semantic conditions unsupported by visual evidence.
\end{itemize}
Outputs failing this structured validation are immediately discarded by the FSM, inherently penalizing the model's mission effectiveness.

\subsection{VLM Variable Scale and Cross-Hardware Deployment Protocol}

To assess the trade-offs associated with deploying vision-language models (VLMs) at the robotic edge, three generative models representing a range of capabilities and parameter sizes were evaluated: Moondream (1.8B), LLaVA-LLaMA3 (8B), and LLaMA 3.2 Vision (11B). In recognition of the significant computational constraints of onboard unmanned aerial vehicle (UAV) systems, the benchmark employs a physical multi-tier hardware setup.

The evaluation utilizes the models across two distinct physical operational environments (Table \ref{tab:hardware_specs}). The High-Performance environment employs an unrestricted desktop computing architecture to establish theoretical performance baselines. In contrast, the Constrained Profile assesses the models natively on a resource-limited mobile GPU architecture. This deliberate physical contrast enables empirical quantification of the architectural requirements necessary to meet the Temporal Reliability Boundary for safe autonomous recovery. to the Temporal Reliability Boundary for safe autonomous recovery.

\begin{table}[h]
\centering
\caption{Physical Computational Profiles Configured for VLM Testing}
\label{tab:hardware_specs}
\resizebox{\columnwidth}{!}{%
\begin{tabular}{|l|c|c|}
\hline
\textbf{Parameter} & \textbf{Profile A: High-Performance Ground Station} & \textbf{Profile B: Constrained Edge Computing} \\ \hline
Processing Target  & Intel Core i5-13600K & Intel Core i7-13650HX \\ \hline
GPU Architecture & NVIDIA RTX 4070 Ti (Desktop) & NVIDIA RTX 4050 (Mobile/Laptop) \\ \hline
VRAM Capacity & 12 GB GDDR6X Dedicated & 6 GB GDDR6 Dedicated \\ \hline
Inference Engine & Ollama (Unrestricted CUDA Pipeline) & Ollama (Constrained CUDA Pipeline) \\ \hline
Evaluation Purpose & Theoretical Fast-Response Baseline & Realistic VRAM-Bottleneck Approximation \\ \hline
\end{tabular}%
}
\end{table}

\subsection{Evaluation Scenarios os Test Evaluation}

\begin{figure}[t]
    \centering
    \includegraphics[width=1.06\columnwidth]{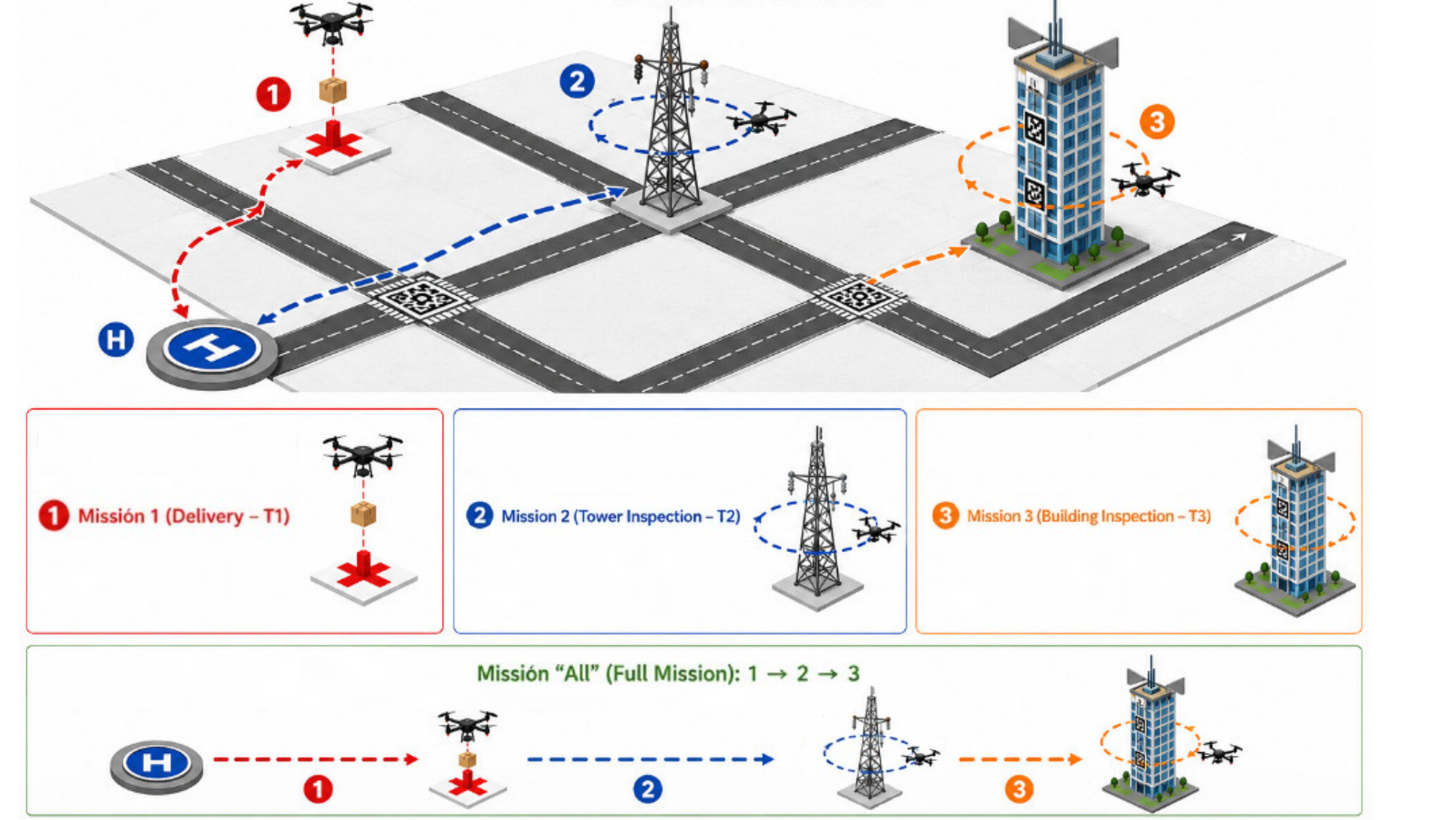}
    \caption{Overview of sequential mission profiles scaling generated for the benchmark.}
    \label{fig:missions}
\end{figure}

The models were deployed across a matrix of incrementally complex operational scenarios simulating autonomous delivery, infrastructure inspection, and Urban Search and Rescue (USAR). The missions range from simple street-tracking (Mission 1) to multi-stage altitude sweeping and dynamic semantic identification across structures (Mission 3). A comprehensive \textit{Full Run} sequential challenge evaluates energy and persistent context capability across all tasks (Figure \ref{fig:missions}).

\begin{figure*}[t]
    \centering

    \begin{minipage}{0.24\textwidth}
        \centering
        \includegraphics[width=\linewidth]{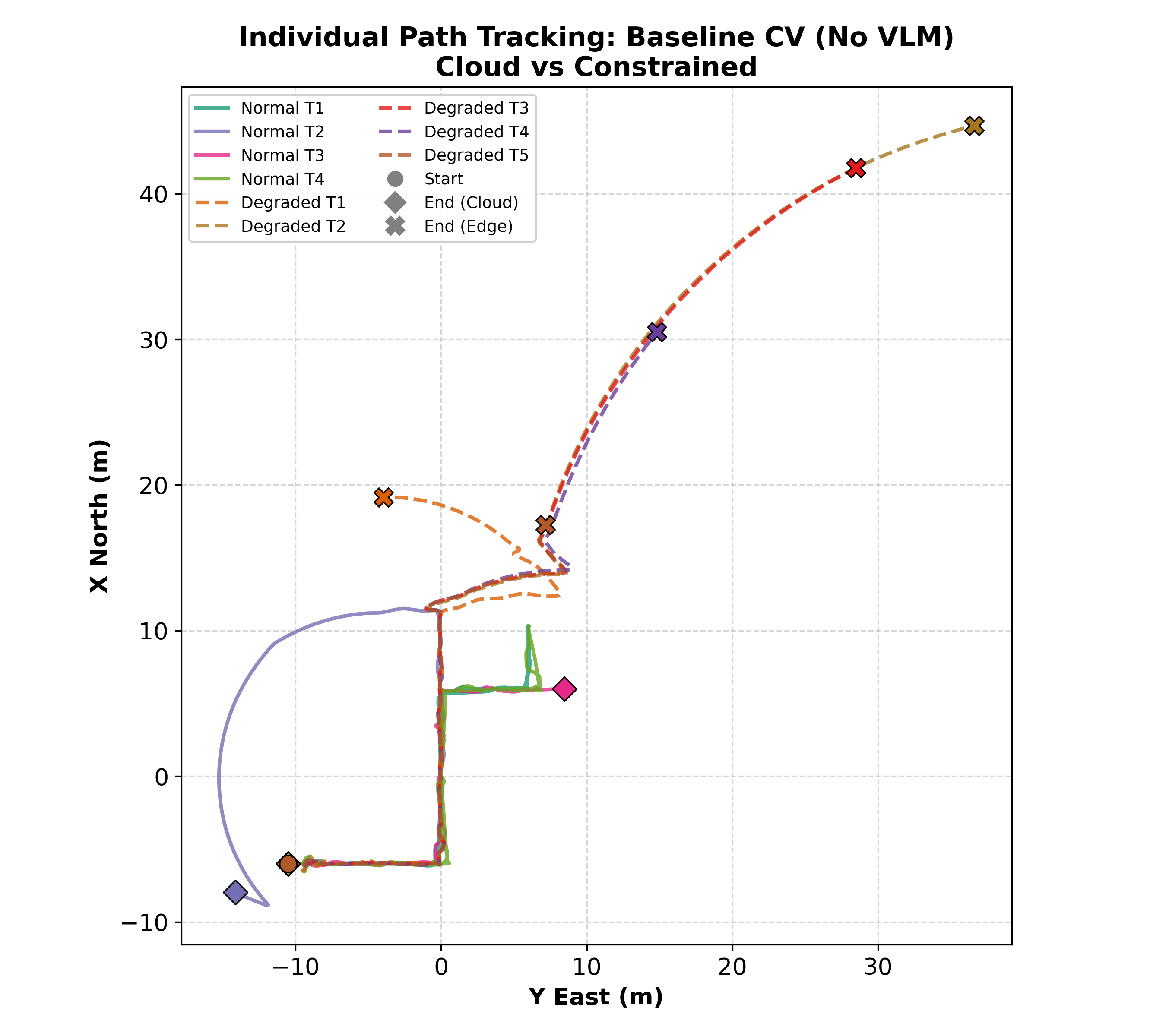}
    \end{minipage}
    \hfill
    \begin{minipage}{0.24\textwidth}
        \centering
        \includegraphics[width=\linewidth]{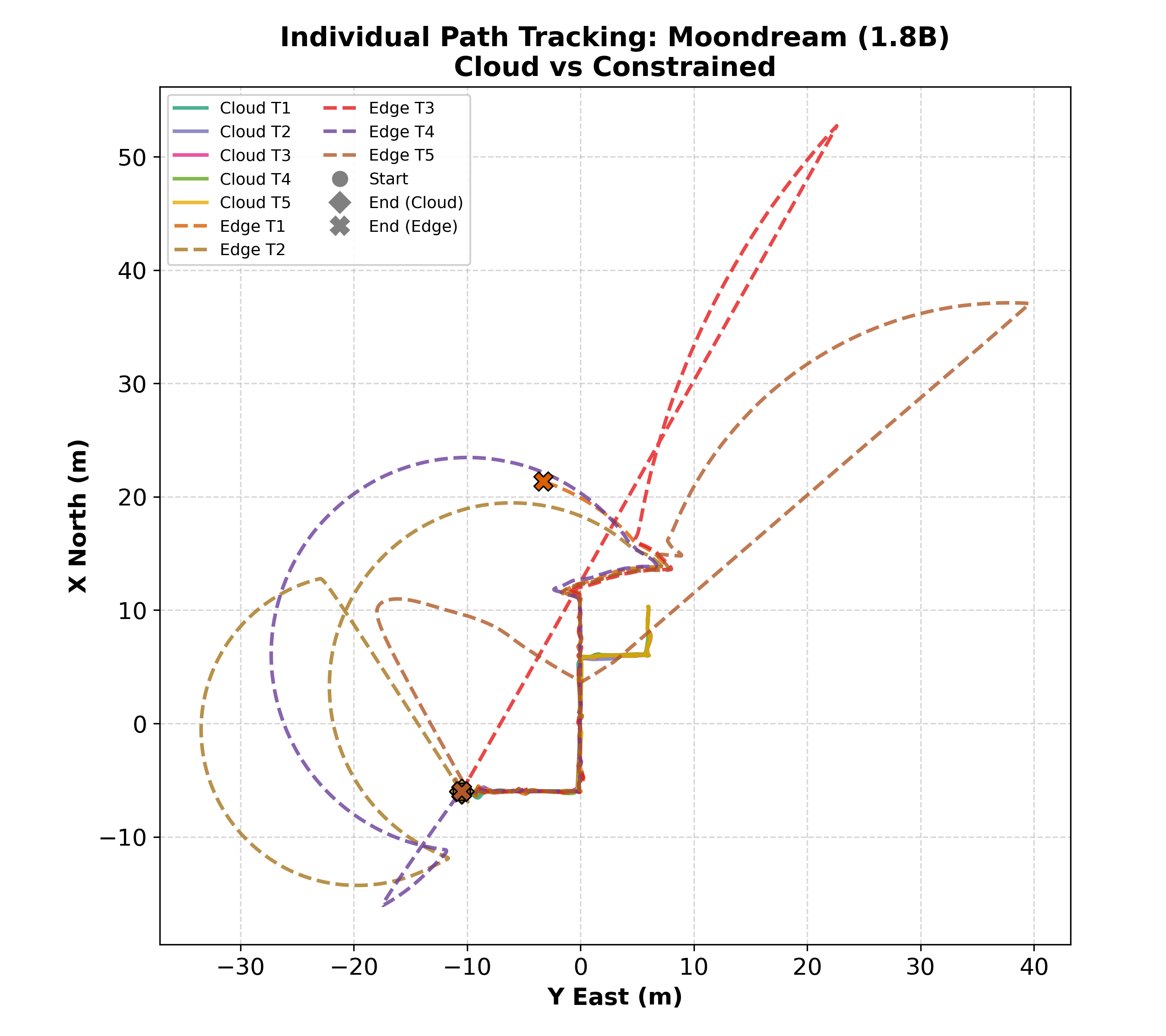}
    \end{minipage}
    \hfill
    \begin{minipage}{0.24\textwidth}
        \centering
        \includegraphics[width=\linewidth]{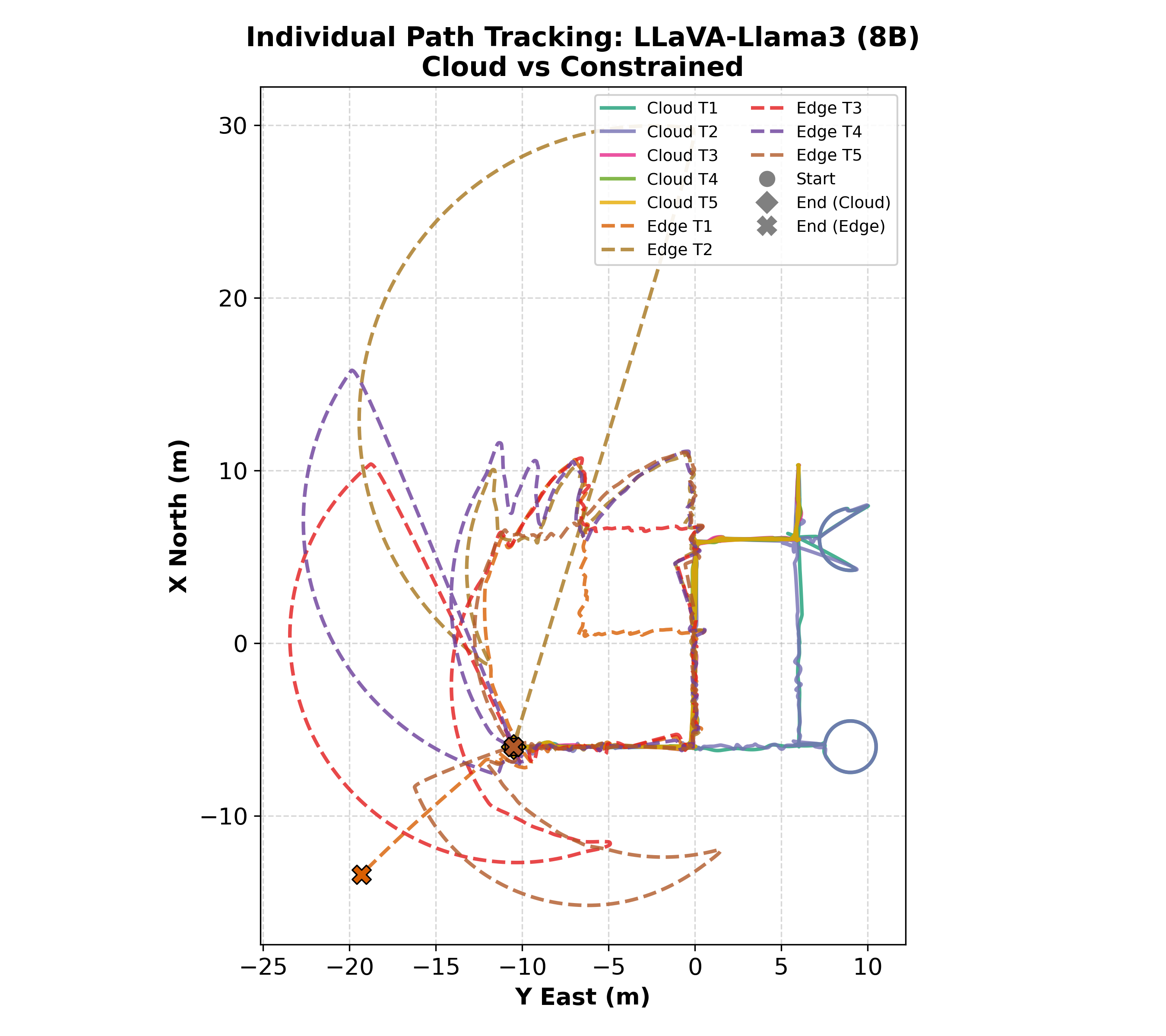}
    \end{minipage}
    \hfill
    \begin{minipage}{0.24\textwidth}
        \centering
        \includegraphics[width=\linewidth]{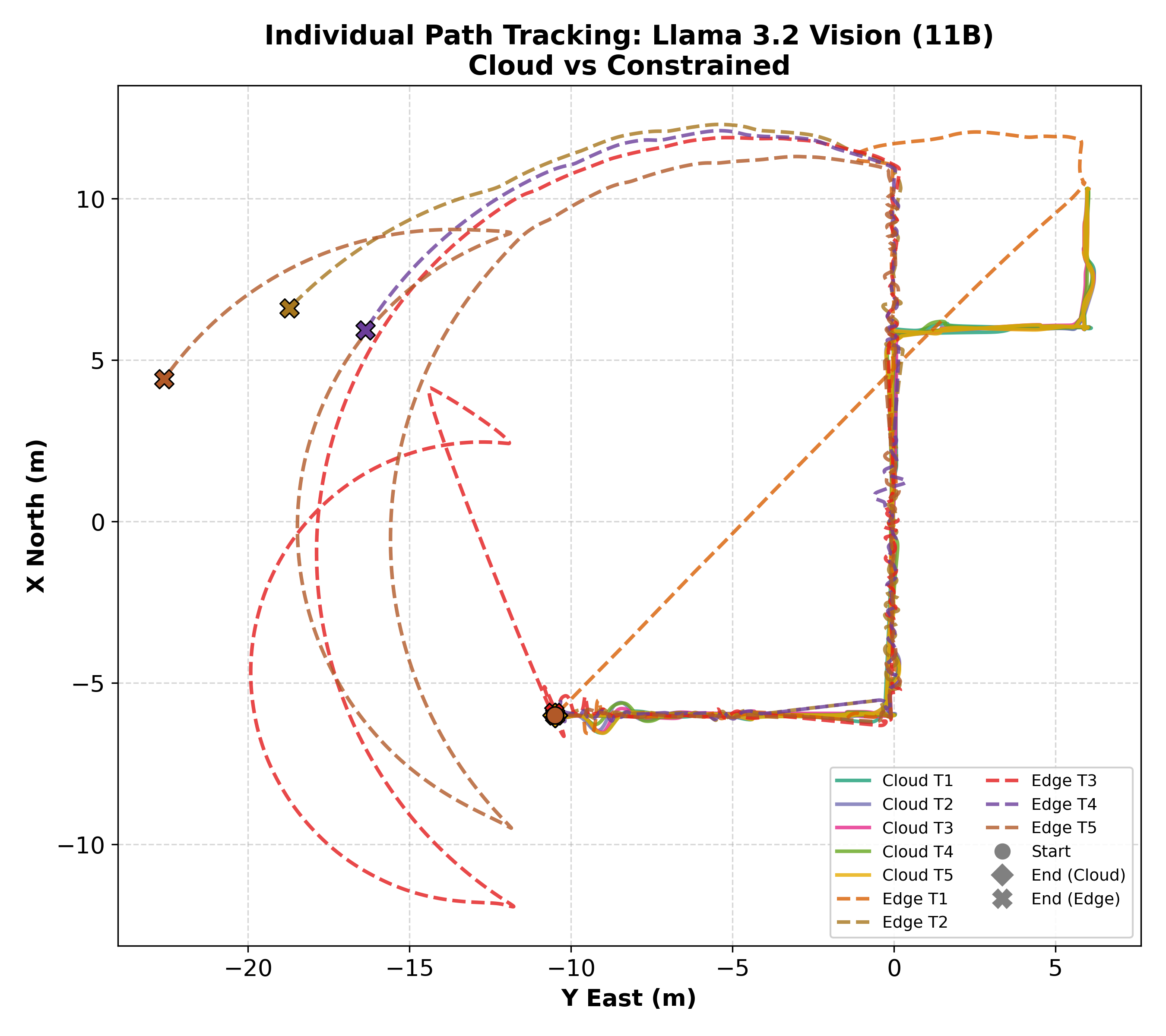}
    \end{minipage}

    \caption{Overlay of trajectories in a stress environment of Mission 1. The baseline model fails at the intersection, whereas the VLM backend attempts to use backtracking to recover the mission path.}
    \label{fig:cross_model_traj}
\end{figure*}

To promote reproducibility and statistical reliability, each
experimental configuration was evaluated using a comprehensive
experimental matrix. Mission~1 was designated as the baseline
challenge because it encompasses the essential trajectory-following
and semantic recovery tasks within a controlled environment. For each Vision-Language Model (VLM) and hardware profile, the
asynchronous semantic query interval was systematically varied across
the set $ \{5,10,15,20,25,30 \} $s. Each stress condition was
independently repeated five times, resulting in 30 autonomous executions per model configuration. 

The effect of computational and communication delays on closed-loop
control performance has been extensively examined in networked
control systems. Previous studies have shown that delays must remain
below system-dependent admissible bounds to preserve control
performance and stability \cite{lian2002network,kim2003maximum}.
This challenge is especially significant in the context of vision-based UAV navigation,
where delays in perception and processing have a direct impact on controller
tuning, trajectory tracking accuracy, and the boundaries of system stability
\cite{humais2022delay}.
The concept of a recoverable region is also consistent with
safety-critical control formulations in which the system state must
remain within a prescribed safe set \cite{taylor2020learning}.

Assuming a nearly constant translational velocity during operation,
and with the finite-state machine providing low-level stabilization,
the displacement accumulated during inference can be approximated by
$e \approx v_{\mathrm{UAV}} t_{\mathrm{inf}}$. Therefore, the temporal
safety limit is defined as follows:

\begin{equation}
TRB=\frac{e_{\mathrm{max}}}{v_{\mathrm{UAV}}},
\label{eq:trb}
\end{equation}

where $e_{\mathrm{max}}$ denotes the maximum admissible spatial
tracking error from which the UAV can recover without exceeding the
environmental boundaries, and $v_{\mathrm{UAV}}$ denotes its average
translational velocity. In the evaluated indoor testbed, the corridor
geometry imposes $e_{\mathrm{max}}\approx2.5$~m. For the nominal
velocity $v_{\mathrm{UAV}}\approx0.5$~m/s, the resulting TRB is

\begin{equation}
TRB=\frac{2.5}{0.5}\approx5.0~\mathrm{s}.
\end{equation}

Stable operation is expected when

\begin{equation}
t_{\mathrm{inf}}<TRB,
\label{eq:trb_stable}
\end{equation}

because the semantic response is obtained before the estimated
displacement exceeds the recoverable region. Conversely,

\begin{equation}
t_{\mathrm{inf}}\geq TRB
\label{eq:trb_unstable}
\end{equation}

This finding indicates that the response latency reaches or exceeds the
first-order recoverability limit. Under these circumstances, delayed
semantic decisions may contribute to trajectory overshoot, collisions,
or mission interruptions.

The proposed Trajectory Recoverability Boundary (TRB) establishes a direct relationship between Vision-Language Model (VLM)
inference latency and the spatial constraints of the robotic
platform. Therefore, it should be interpreted as a platform- and
scenario-specific criterion that can be applied when comparing
different VLM architectures and inference hardware.

\section{Results and Discussion}

\subsection{Performance and Inference Dynamics Across Hardware Profiles}

The initial analysis establishes a foundation for assessing whether the scale of the VLM architecture imposes fundamental limitations on real-time stability under stringent hardware constraints. To determine these thresholds, benchmarks were conducted on both a dedicated GPU workstation (High-Performance Profile) and a mobile UAV-grade system (Constrained Edge Profile). Each configuration underwent incremental evaluation through multi-trial iterations under varied query conditions. A trial was deemed successful if the UAV completed the mission autonomously, without experiencing structural hallucination deadlocks or kinetic collisions resulting from delayed processing.

\begin{figure}[h]
    \centering
    \includegraphics[width=0.99\linewidth]{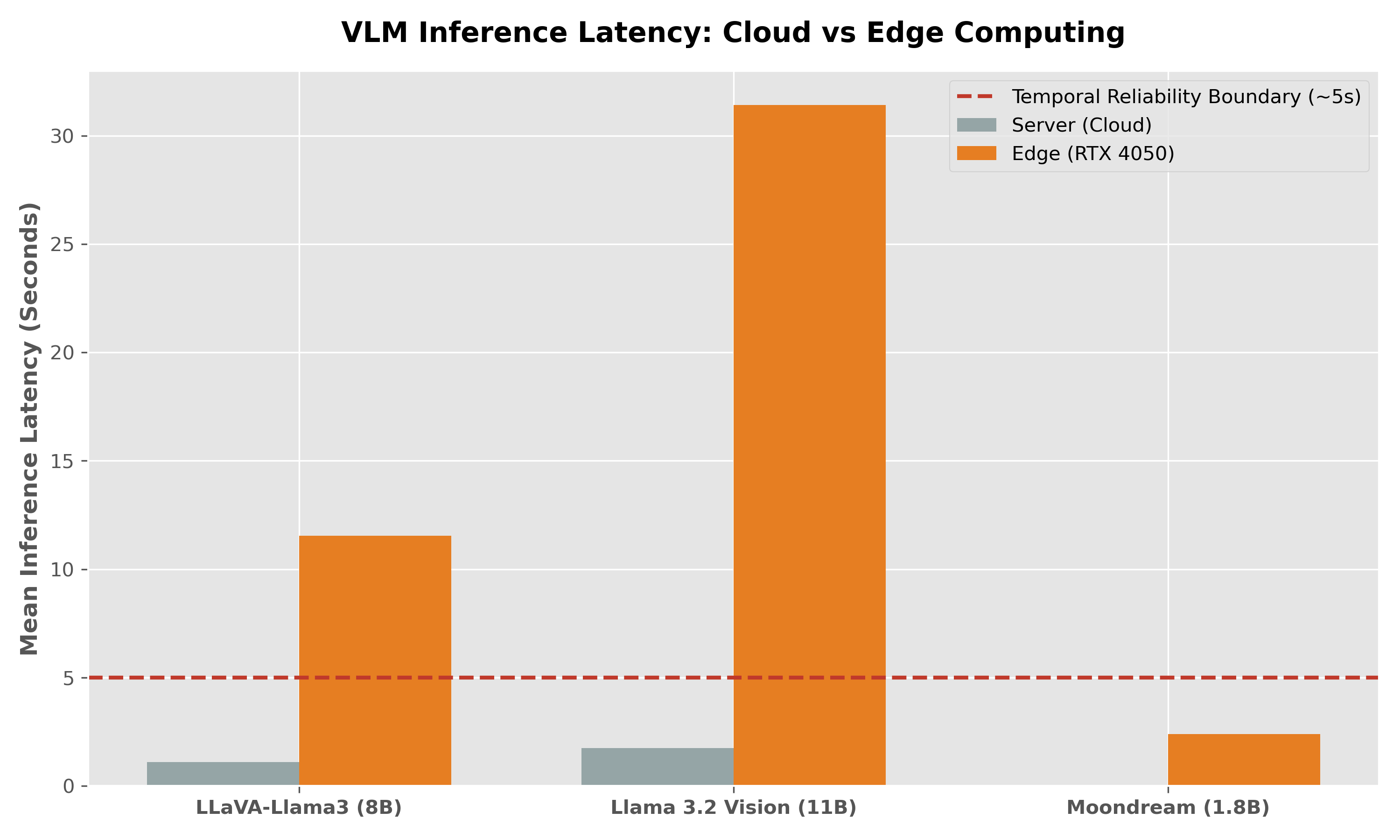} 
    \vspace{-0.15cm}
    \caption{VLM Inference Latency Comparison between Cloud (RTX 4070 Ti) and Edge (RTX 4050) computing profiles. The red dashed line denotes safety threshold of the Temporal Reliability Boundary.}
\label{fig:latency_distribution}
    \vspace{-0.25cm}
\end{figure}

Prior to assessing physical trajectory deviations, it is necessary to quantify the temporal overhead resulting from the varying parameter scales of integrated vision-language models (VLMs). Figure \ref{fig:latency_distribution} presents the average inference latencies that the control loop (finite state machine, FSM) must accommodate asynchronously during mission execution. By comparing these delays across both cloud and edge computing profiles, the graph reveals significant computational scaling bottlenecks that impact robotic flight stability.

The empirical distribution in Figure~\ref{fig:latency_distribution} demonstrates that the parameter-to-latency ratio incurs a pronounced non-linear scaling penalty when shifting from an unrestricted cloud environment to a power-constrained edge architecture. In the High-Performance Profile, dedicated GPU tensor cores efficiently manage the computational demands of the large 11B-parameter model, maintaining the average inference cycle at approximately $3.5$ seconds. Under these conditions, the asynchronous FSM effectively bridges semantic gaps and preserves the drone's spatial stability.

In contrast, implementation on the restricted edge profile highlights the instability associated with deploying model scales not optimized for robotics. The highly quantized Moondream (1.8B) model remains robust, processing queries with an average latency of 2.3 seconds, which is well within the critical safety margin. However, the similarly sized LLaMA 3.2 Vision (11B) model experiences significant architectural degradation. Lacking extensive parallel processing and sufficient VRAM, the 11B model exhibits an average inference latency of 31 seconds per query. This deterioration is primarily due to VRAM scarcity, which forces resource-constrained systems to engage in continuous memory offloading and pooling on the host's slowest bus.

These computational constraints ensure that the robotic platform will exceed the temporal reliability limit, as indicated by the red limit constraint. When edge latencies surpass thirty seconds per directional decision, the FSM operates without spatial updates for durations incompatible with the natural drift inertia of unmanned aerial vehicles in indoor environments. Even with optimal sensor fusion calibration, the UAV inevitably exits the safety zone, demonstrating that non-quantized monolithic vision-language models are fundamentally incompatible with the temporal requirements of autonomous edge robotics.

\subsection{Discussion on Resilience and The Temporal Boundary}

High-fidelity SITL validation using PX4 and Gazebo confirms that deploying large generative models completely paralyzes the navigation loop when shifted to resource-constrained environments. As detailed in Table \ref{tab:vlm_navigation_performance}, while the LLaMA 3.2 (11B) copilot provides absolute flawless capabilities on the Cloud ($100.0\%$ success), this same architecture collapses under the strain of the Edge GPU, suffocated by VRAM shortages that inflate the mean latency to $31.4$s. 

\begin{figure}[h]
    \centering
    \includegraphics[width=1.0\columnwidth]{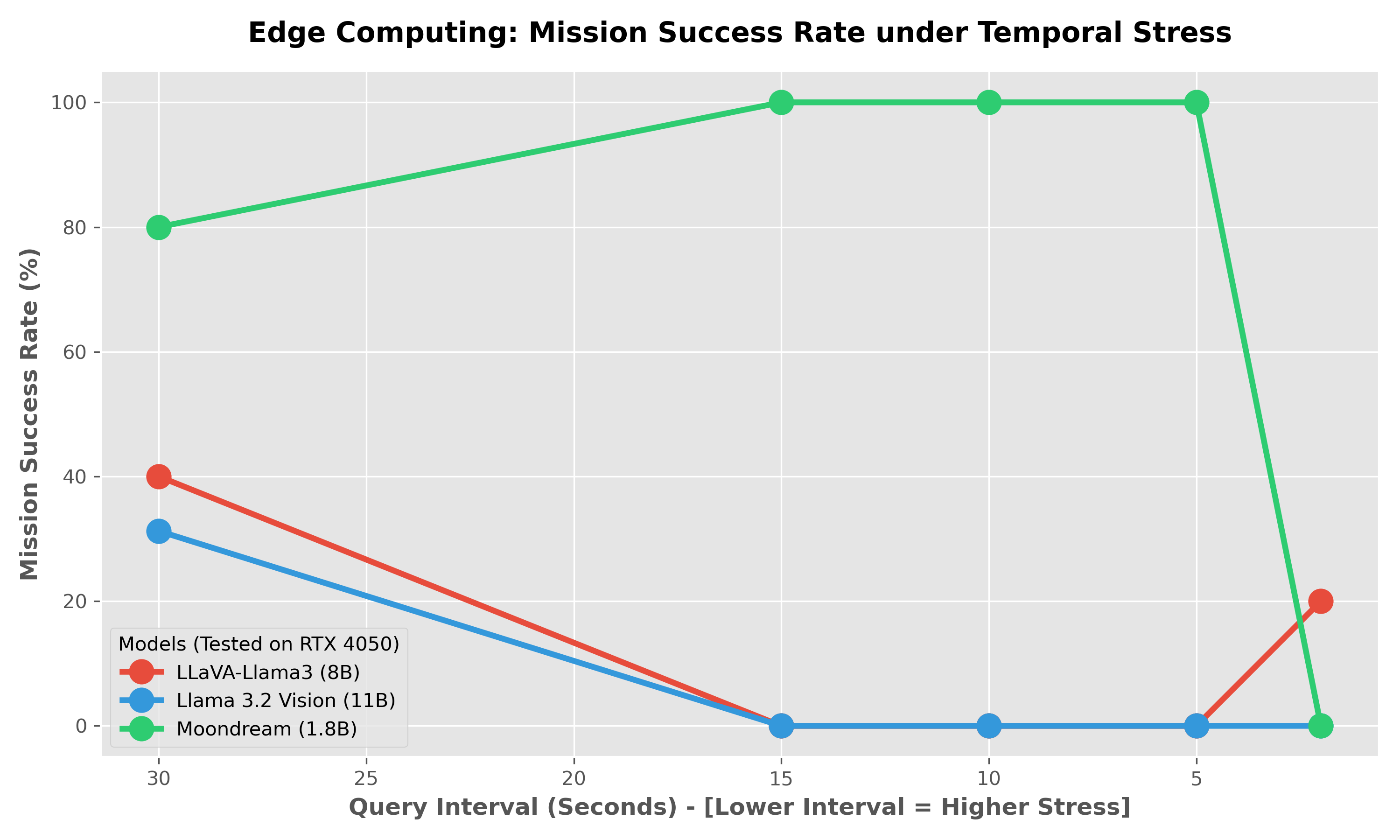} 
    \caption{Temporal Reliability Boundary Isolation: The destructive impact of aggressive Query Frequencies on physical flight stability on Edge hardware (RTX 4050).}
    \label{fig:stress_test_freq}
\end{figure}

\begin{table*}[t]
\centering
\caption{Comprehensive comparison of VLM-based autonomous navigation across Cloud and Edge architectures under nominal stress. Success Rate tracks mission completion without hardware-induced FSM timeouts.}
\label{tab:vlm_navigation_performance}
\small
\setlength{\tabcolsep}{10pt}
\begin{tabular}{llcccc}
\hline
Hardware Profile & Inference Model & Mean Latency (s) & Max Error (m) & Format Stability & Success Rate \\
\hline
\multirow{3}{*}{\rotatebox[origin=c]{90}{Desktop}}
& Moondream (1.8B)    & 0.702 & 0.798 & 14.5\% & 95.0\% \\
& LLaVA-Llama3 (8B)   & 3.284 & 0.798 & 99.5\% & 100.0\% \\
& LLaMA 3.2 (11B)     & 3.312 & 0.804 & \textbf{100.0\%} & \textbf{100.0\%} \\
\hline
\multirow{3}{*}{\rotatebox[origin=c]{90}{Laptop}}
& Moondream (1.8B)    & \textbf{2.394} & \textbf{0.751} & 16.5\% & \textbf{80.0\%} \\
& LLaVA-Llama3 (8B)   & 11.538 & 14.56 & 99.2\% & 40.0\% \\
& LLaMA 3.2 (11B)     & 31.413 & 35.19 & \textbf{100.0\%} & 31.2\% \\
\hline
\end{tabular}
\end{table*}

The quantized Moondream (1.8B) model efficiently stabilizes the system, achieving an $80.0\%$ baseline mission success rate despite significant edge hardware constraints. This performance is enabled by maintaining low inference latencies, averaging $2.3$ seconds per query. This empirical contrast provides physical validation for the proposed asynchronous FSM-VLM control architecture. The results indicate that the fast-reacting, low-level FSM loop robustly safeguards fundamental flight stability, while the Vision-Language Model independently resolves high-level semantic ambiguities through latency-aware trajectory backtracking. However, this stability depends on a critical equilibrium: the generative engine must complete its semantic inference before the UAV's kinetic inertia carries the vehicle beyond the point of safe recovery, a threshold defined as the Temporal Reliability Boundary.

Figure~\ref{fig:stress_test_freq} illustrates the degradation in mission success on Edge Computing hardware across all models. Implementing an aggressive visual interaction loop exposes and ultimately breaches the Temporal Reliability Boundary. The mean inference latency of the 11B-parameter architecture exceeds the $31$-second threshold on the GPU, rendering the hardware incapable of real-time multi-query processing. As a result, submitting a visual query every $5.0$ or $10.0$ seconds saturates the asynchronous pipeline.

These empirical findings provide a foundation for future deployment of embodied AI. The results demonstrate that in computationally constrained autonomous environments, semantic capability must be balanced against execution speed, as temporal latency is the primary determinant of system survival. The evidence indicates that catastrophic, hardware-induced robotic collisions can only be systematically avoided by reducing the generative model's parameter footprint to fit within the robotics’ real-time inference threshold.

Geometric divergence becomes catastrophic when heavier VLM architectures are evaluated on edge hardware. While the LLaVA (8B) and LLaMA (11B) models generate continuous mission paths with unrestricted cloud resources, their edge-constrained versions (Panels C and D) exhibit severe spatial dropouts. These are observed as premature collision terminations, visually corroborating the previously discussed mathematical collapse. The significant temporal delay causes the drone to enter localized hovering loops until positional drift overwhelms the sensors, resulting in collisions before a semantic rescue vector can be received. These spatial mappings demonstrate that an architecture's parameter weights influence not only raw intelligence but also the geometric survivability of the physical platform.

\section{Conclusions}

This paper introduces a framework for integrating Vision-Language Models (VLMs) as asynchronous autonomous copilots to support UAV navigation in perceptually and computationally demanding environments. Rather than focusing solely on optimizing model architectures for semantic accuracy, this study explicitly identifies the system's Temporal Reliability Limit. By comparing locally executed VLM backends with a classical Computer Vision (CV) baseline across various hardware configurations, the research quantifies the critical relationship between inference latency and real-time kinetic flight stability.

Validation within the simulation environment demonstrated a significant paradigm shift concerning model scale at the robotic edge. The large-scale LLaMA 3.2 Vision (11B) model exhibited strong semantic reasoning capabilities under unrestricted Cloud conditions but failed under Edge emulation constraints, achieving only a $31.2\%$ success rate due to substantial $31$-second inference delays that compromised the UAV's spatial stability. In contrast, the heavily quantized Moondream (1.8B) model proved to be the most robust operational configuration for Edge architectures. With rapid $2.3$-second query latencies, Moondream maintained the temporal boundary and achieved an $80.0\%$ mission success rate. The classical CV baseline, by comparison, failed completely ($0.0\%$ success) under visual stress constraints in the absence of semantic support.

Future research should focus on expanding and ultimately overcoming the established Temporal Reliability Boundaries through comprehensive, hardware-aware infrastructure optimizations. While this study demonstrates the effectiveness of quantized models at the edge, promising directions include the use of edge-native inference acceleration, such as dedicated Neural Processing Units (NPUs), TensorRT quantization pipelines, and sparse attention mechanisms, to decouple parameter count from execution latency. Additionally, subsequent evaluations will transition these constrained VLM architectures from static testbeds to dynamic, unstructured physical environments. In these scenarios, the VLM generative pipeline will be required to advance beyond binary spatial parsing and deliver complex, predictive semantic reasoning for moving subjects, rapidly changing obstacles, and adversarial visual conditions.

\section*{Acknowledgment}

The authors thank the Robotics and Artificial Intelligence Laboratory of the Technological University of Uruguay for their support of this work.

\bibliographystyle{./bibliography/IEEEtran}
\bibliography{./bibliography/IEEEabrv,./bibliography/main}

\end{document}